\documentclass[letterpaper]{article} 
\usepackage[submission]{aaai24}  
\usepackage{times}  
\usepackage{helvet}  
\usepackage{courier}  
\usepackage[hyphens]{url}  
\usepackage{graphicx} 
\usepackage{natbib}  
\usepackage{caption} 
\usepackage{amsmath}
\usepackage{amssymb}
\usepackage{mathtools}
\usepackage{amsthm}
\usepackage{xcolor}   
\usepackage{bm}
\usepackage{algorithm}
\usepackage{algorithmic}
\usepackage{xspace}

\renewcommand{\[}{\begin{eqnarray}}
\renewcommand{\]}{\end{eqnarray}}

\newcommand\norm[1]{\left\lVert#1\right\rVert}
\usepackage{booktabs}

\DeclareMathOperator{\E}{\mathbb{E}}
\DeclareMathOperator{\R}{\mathbb{R}}

\usepackage{newfloat}
\usepackage{listings}
\DeclareCaptionStyle{ruled}{labelfont=normalfont,labelsep=colon,strut=off} 
\floatstyle{ruled}
\newfloat{listing}{tb}{lst}{}
\floatname{listing}{Listing}
\usepackage{bibentry}

\newcommand{\model}{\textsc{TempoTokens}\xspace}

\newcommand{\indicator}{\bm{1}}

\title{Diverse and Aligned Audio-to-Video Generation via \\ Text-to-Video Model Adaptation}

\author{Guy Yariv $^{\heartsuit, \clubsuit}$, Itai Gat $^\diamondsuit$, Sagie Benaim $^\heartsuit$, Lior Wolf $^\spadesuit$, Idan Schwartz $^{\spadesuit, \clubsuit, *}$, Yossi Adi $^{\heartsuit,}$\thanks{Equal Contribution.}}

\affiliations{
  $^\heartsuit$The Hebrew University of Jerusalem , $^\diamondsuit$Technion  \\ $^\spadesuit$Tel-Aviv University,  $^\clubsuit$NetApp}

\begin{document}
\maketitle

\begin{abstract}
We consider the task of generating diverse and realistic videos guided by natural audio samples from a wide variety of semantic classes. For this task, the videos are required to be aligned both globally and temporally with the input audio: globally, the input audio is semantically associated with the entire output video, and temporally, each segment of the input audio is associated with a corresponding segment of that video. We utilize an existing text-conditioned video generation model and a pre-trained audio encoder model. The proposed method is based on a lightweight adaptor network, which learns to map the audio-based representation to the input representation expected by the text-to-video generation model. As such, it also enables video generation conditioned on text, audio, and, for the first time as far as we can ascertain, on both text and audio. We validate our method extensively on three datasets demonstrating significant semantic diversity of audio-video samples and further propose a novel evaluation metric (AV-Align) to assess the alignment of generated videos with input audio samples. AV-Align is based on the detection and comparison of energy peaks in both modalities. In comparison to recent state-of-the-art approaches, our method generates videos that are better aligned with the input sound, both with respect to content and temporal axis. We also show that videos produced by our method present higher visual quality and are more diverse. Code and samples are available at: \url{https://pages.cs.huji.ac.il/adiyoss-lab/TempoTokens}.
\end{abstract}

\section{Introduction}
Neural generative models have changed the way we create and consume digital content. From generating high-quality images and videos~\citep{ho2020denoising, rombach2022high}, speech and audio~\citep{wang2023neural, sheffer2023hear, copet2023simple, kreuk2022audiogen, hassid2023textually}, through generating long textual spans~\citep{touvron2023llama, touvron2023llama2, brown2020language}, these models have shown impressive results. 

In the context of video generation, progress has been more elusive, with recent work making progress in generating short videos conditioned on text~\citep{singer2022make, ho2022imagen}. Although audio is tightly connected to videos (e.g., providing important cues for motion in a scene), most of the prior work did not consider audio in the generation process. For instance, the action of `playing drums' or the `motion of waves' can be distinctively associated with a naturally occurring sound. Moreover, audio is comprised of structural components such as pitch and envelope that provide important cues for the type of scene and motion depicted.

\begin{figure}[t!]
    \centering
    \includegraphics[width=\linewidth]{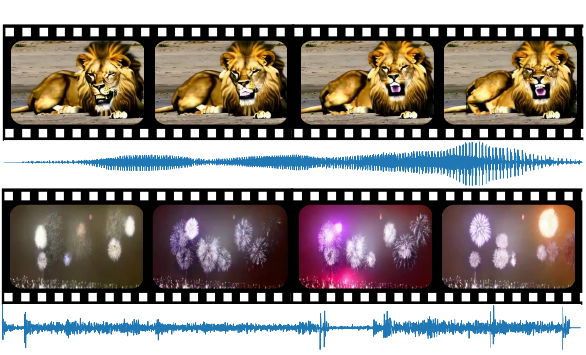}
    \caption{Generated video frames (above) and input audio signal (below the frames) employing our technique. The input to our model is an audio recording from which a representation is extracted. This representation maintains crucial temporal attributes and is then mapped into a text-based latent space representation incorporating both local and global audio context. Subsequently, this latent representation is fed into a pre-trained text-to-video diffusion generative model, ensuring the synchronized generation of video which is closely aligned with the input audio.}
    \label{fig:teasher_fig}
\end{figure}

We tackle the problem of generating diverse and realistic videos guided by natural audio samples. Our generated videos capture diverse and real-life settings from a wide variety of semantic classes and are aligned both globally and temporally with the input audio. Globally, the input audio is semantically associated with the entire output video, and temporally, each segment of the input audio is associated with a corresponding segment of that video. An example generation video can be seen in Figure~\ref{fig:teasher_fig}.

Prior work on audio-guided video generation was mainly focused on either global information in the videos (i.e., capturing the semantic class) or specific scenes (e.g., speech). \citet{mama2021nwt, park2022synctalkface, kumar2020robust} generate talking heads conditioned on speech, but these are limited to videos of human faces and are conditioned on speech and not natural audio. More closely related to our setting, given an input video and an audio sample,  \citet{chatterjee2020sound2sight} generate a continuation of the video that is aligned with the audio. Our method, however, generates videos from audio-only. \citet{ge2022long} proposed a method for generating aligned videos conditioned on audio. While impressive, generated videos are highly limited in diversity. 
Other works such as \citep{chen2017deep, hao2022vag, ruan2023mm} generate videos that are globally aligned to the semantic class of the input audio sample (e.g.,  dancing, drums, etc.) but are unable to generate videos in which every segment is temporally aligned to each segment in the input audio sample. 

In contrast to the above methods, our approach enables the generation of diverse and realistic videos associated and aligned with the input audio from a wide variety of semantic classes. Our work utilizes a pre-trained text-conditioned video generation engine and converts the input audio to a sequence of pseudo tokens. Given an input audio sample, we first encode it using an audio encoder, producing a spectral representation of the audio signal. To capture local-to-global information, we construct the representation considering the $i$-th segment as well as neighboring segments. In particular, we use windows of varying sizes and average the embeddings corresponding to audio segments in these windows. Next, to produce the $N$-th video frame, we divide the audio embedding into $N$ consecutive  segments. We then train an adapter network to map each of these segments to a set of pseudo-tokens. Lastly, to produce the corresponding video, we feed the output of the audio mapping module into the pretrained text-to-video generation model.

Intuitively, we learn a mapping between the audio representation obtained by the pre-trained audio encoder, to the textual tokens' representation used for conditioning the pre-trained text-to-video model. By that, extending the possible video conditioning to audio tokens. To validate our approach, we consider a number of datasets that exhibit a diverse set of videos and input audio samples. We consider the Landscape dataset~\citep{lee2022sound}, which captures landscape videos. The AudioSet-Drums dataset~\citep{gemmeke2017audio} which captures drums videos, and the VGGSound dataset~\citep{chen2020vggsound} which consists of a diverse set of real-world videos from 309 different semantic classes. 

We compare our method to state-of-the-art approaches, both in terms of objective evaluation and human study. We evaluate the audio-video alignment as well as video quality and diversity. To capture temporal alignment, we devise a new metric based on detecting energy peaks in both modalities separately and measuring their alignment. Further, we provide an ablation study where we consider alternative approaches to condition the video model. 

\noindent{\textbf{Our contributions:}} (i) A state-of-the-art audio-to-video generation model which captures diverse and naturally occurring real-life settings from a wide variety of input videos of different semantic classes; (ii) We present a method that is based on a lightweight adapter, which learns to map audio-based tokens to pseudo-text tokens. As such, it also allows video generation conditioned on text, audio, or both text and audio. As far as we are aware, our method is the first to enable video generation conditioned both on audio and text; and (iii) Our method can generate natural videos aligned with the input sound, both globally and temporally. To validate this, we present a novel evaluation function to measure audio-video alignment. Since, as far as we can ascertain, we are the first to generate diverse and natural videos guided by audio inputs, such an evaluation function is critical to making progress in the field.
\begin{figure*}[t!]
    \centering
    \includegraphics[width=0.74\linewidth]{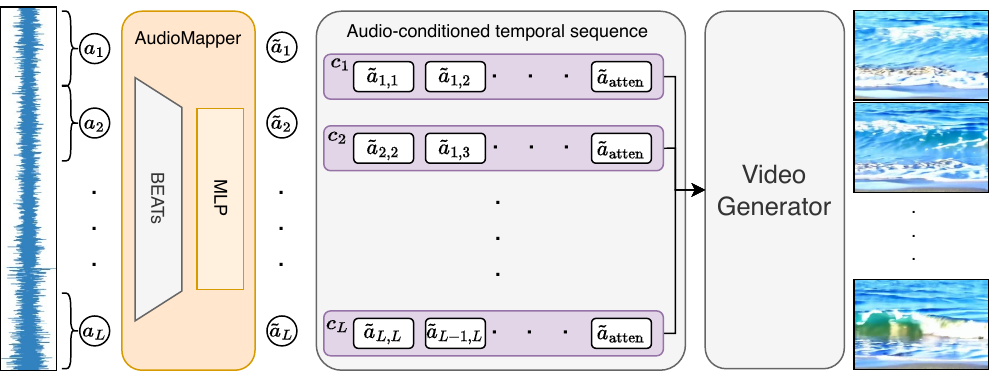}
    \caption{An illustration  of the proposed model architecture and method. The input audio is first passed through a pre-trained audio encoder model (BEATs). Then, the resulting  representations are fed into a trainable MLP layer, establishing a mapping between audio and text tokens. These text-based representations are then used to condition each frame via a temporal audio-conditioned sequence. This sequence effectively takes into account both local and global audio segments. Furthermore, an attentive token ($\tilde a_{\text{atten}}$) is included to learn the identification of significant audio signals using a pooling attention layer. Lastly, the conditioned components are utilized to generate frames through a pre-trained video generator. Notably, optimization is only applied to the MLP within the AudioMapper model and the pooling attention module.}\label{fig:arch}
\end{figure*}

\section{Related work}

\noindent{\textbf{Audio-to-image generation.}} 
Text-to-image generation has seen great advances recently, using either autoregressive methods~\citep{dalle,gafni2022make,parti} or diffusion based models~\citep{nichol2021glide,ramesh2022hierarchical,imagen,rombach2022high, ramesh2022hierarchical, rombach2022high}. This inspired a new line of work concerning audio-to-image generation. \citet{zelaszczyk2022audio,wan2019towards} proposed to generate images based on audio recordings using a GAN.~\citet{zelaszczyk2022audio} present results for generating MNIST digits only and did not generalize to general audio sounds, while \citet{wan2019towards} generate images from general audio. In Wav2Clip~\citep{wu2022wav2clip}, the authors learn a Contrastive Language-Image Pre-Training (CLIP)~\citep{radford2021learning} like a model for learning joint representation for audio-image pairs. Later on, such representation can be used to generate images using VQ-GAN~\citep{esser2021taming} under the VQ-GAN CLIP~\citep{crowson2022vqgan} framework. The most relevant related work to ours is AudioToken~\citep{yariv2023audiotoken}, in which the authors learn an audio token while adapting a diffusion-based text-to-image model to generate images using audio inputs. 

\noindent{\textbf{Text-to-video generation.}} Early attempts to establish a connection between text and video relied on conditioned retrieval methods~\citep{ali2022video}. Later, \citet{wu2021godiva} introduces the novel integration of 2D VQVAE and sparse attention in text-to-video generation, facilitating the generation of highly realistic scenes. ~\citet{wu2022nuwa} extends this method and presents a unified representation for various generation tasks in a multitask learning scheme. Later on, CogVideo \citep{hong2022cogvideo} is built on top of a frozen text-to-image model by adding additional temporal attention modules. \citet{singer2022make} further improves generation quality following a similar modeling paradigm. Video Diffusion Models~\citep{he2022latent} uses a space-time factorized U-Net with joint image and video data training. 
Other approaches, such as \citet{villegas2022phenaki} and 
\citet{villegas2022phenaki} and ~\citet{yu2023magvit} proposed transformer-based approaches to generate long videos or for multi-task-learning. 
The most relevant prior work to ours is \citet{wang2023modelscope}, which proposed ModelScope. ModelScope is a latent diffusion-based text-to-video generation model with spatiotemporal blocks. By that, ModelScope enables consistent frame generation and smooth movement transitions.  

\noindent{\textbf{Audio-to-video generation}} models can be roughly divided into two: (i) speech-to-video generation (talking heads); and (ii) general audio-to-video. Under the speech-to-video generation, \citet{mama2021nwt} proposed learning a discrete latent representation of the video signal using VQ-VAE, which will be later modeled via an auto-encoder conditioned on speech spectrogram. \citet{park2022synctalkface} generates talking face focusing a piece of phonetic information via \emph{Audio-Lip Memory} module, while \citet{kumar2020robust} proposed a one-shot approach for fast speaker adaptation. 

When considering general audio-to-video generation, \citet{chatterjee2020sound2sight} first proposed a method of generating aligned videos conditioned on both audio and video prompts. \citet{ge2022long} introduced a transformer-based approach for generating videos conditioned on either audio or textual features. Although providing impressive generations, their videos are not diverse and were demonstrated on drum generation only. \citet{chen2017deep} suggest using separate frameworks for audio-to-image and image-to-audio generation. \citet{hao2022vag} also suggest modeling both audio-to-image and image-to-audio using bidirectional transformers, however, using a unified framework. The authors prove it is better than two separate ones. Lastly, \citet{ruan2023mm}, follows the same modeling paradigm, however, using latent diffusion models.
\section{Method}
The proposed method is composed of three main components: (i) an AudioMapper; (ii) multiple audio-conditioned temporal sequences; and (iii) a text-to-video generation module. As our goal in this study is to enrich video generation models using audio inputs, we leverage a pre-trained diffusion-based text-to-video model and augment it with audio conditioning capabilities. A visual description of the proposed method can be seen in Figure~\ref{fig:arch}.

In contrast to converting audio to image, transforming audio to video presents two additional challenges: (i) ensuring the creation of coherent frames and (ii) synchronization between the audio and video components. For example, consider the scenario of having an audio recording of a dog barking. In the resulting video, it is crucial not only for the dog's appearance to remain consistent across all frames but also for the match between the timing of the barking sound and the dog's motion. In this work, we focus on item (ii) by temporally conditioning the generation of each of the video frames by a contextualized representation of the input audio. 

Formally, we are interested in the generation of a video, denoted as $v=(v^{(1)},\ldots,v^{(L)})$, where $v^{(i)} \in \mathbb{R}^{3\times H \times W}$ is an output frame, driven by a corresponding audio condition $a=(a_1,\ldots, a_{R})$, where $a_i \in [-1, 1]$ is an audio sample at a given sampling rate in the time domain. We seek to establish a conditional probabilistic model, $p_{\theta}(v|a)$,  encompassing the entire frame-set, where each frame $v^{(i)}$ is conditioned on $a$, which denotes the audio condition. 

Note that the conditioning of each frame considers the entire audio input but is built differently for each frame. More details can be found in the paragraph on Audio-conditioned temporal sequence. 

\noindent{\textbf{AudioMapper}}
 maps the audio representation obtained from a pre-trained audio encoder to pseudo-tokens compatible with the pre-trained text-to-video model. We denote the output of the AudioMapper as \model. 

Formally, the model gets as input embedded audio, which originates from a pre-trained audio encoder $h: [-1, 1]^{R}\rightarrow \mathbb{R}^{R'\times H\times d}$, where $H$ is the number of layers the representation is collected from, $d$ is the inner dimension of the encoder, and $R'$ is the segment length that $h$ operates on. To force both audio and video latent representations to have the same dimension, we fix $R'=L$ by employing a pooling layer. Specifically, we use the BEATs model~\cite{chen2022beats} as the audio encoder $h$. Different layers encapsulate a range of specificity levels. Representations derived from BEATs' final layers are strongly tied to class-related attributes, whereas earlier layers encompass low-level audio features~\cite{gat2022latent, adi2019reverse}. We embed an audio segment into a token representation using a non-linear neural network $g: \mathbb{R}^{L\times H\times d}\rightarrow\mathbb{R}^{L\times  H\times d_t}$:
\[
    \tilde a^{(i)} = g \left( h(a)^{(i)} \right),
\]
where $\tilde a^{(i)} \in\mathbb{R}^{L\times H \times d_t}$, and $d_t$ is the embedding dimension of the text-conditioned tokens of the video generation process. The network $g$ consists of four sequential linear layers with GELU non-linearity between them. We denote $\tilde a^{(i)}$ as \model. Subsequently, we generate a temporal conditioning sequence for each video frame using \model. We provide a detailed description of the process in the following paragraph.

\begin{figure}[t!]
    \centering
    \includegraphics[width=1\linewidth]{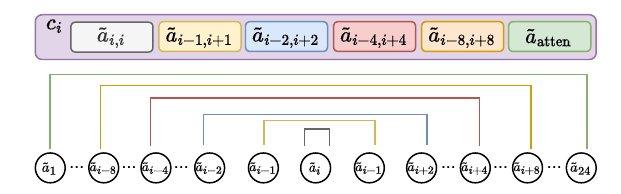}
    \caption{Illustration of the audio-conditioned temporal sequence for the case of 24 audio components. For the $i$-th frame,  the window sizes grow exponentially,  considering local audio details to aid in aligning audio and video, as well as the broader global information that enhances the differentiation of video classes. Additionally, we introduce a token that encompasses all audio components and identifies the significant ones through an attention pooling layer ($\tilde a_{\text{atten}}$).}
    \label{fig:tempo_sequence}
\end{figure}

\noindent{\textbf{Audio-conditioned temporal sequence.}} Next, to better capture the local context around each video frame, we apply an expanding \emph{context window} technique over the obtained \model. This approach captures the surrounding sound signals of the $i$-th frame as follows:
\[
    c^{(i)} = \left(\tilde a_{\max(1, i-j),\min(i+j, K)} \mid j = 2^k \right)_{k=0}^{\log{K}},
\]
where
\[
    \tilde a_{l,r}=\frac{1}{r-l}\sum_{s=l}^{r}\tilde a^{(s)}.
\]
This context window expands exponentially with increasing temporal distance from the target position, facilitating consideration of a wider local-to-global audio context range. The exponential expansion effectively balances local and global contexts, encompassing important distant audio components that can provide valuable insights into the audio class and close temporal changes needed for audio-video alignment. Figure~\ref{fig:tempo_sequence} visually describes the audio-conditioned temporal sequence.  
Finally, we consider a context window that encompasses all audio signals. We substitute average operation with a trainable attentive pooling layer~\cite{schwartz2019factor}. Thus,
\[
    \tilde   a_{\text{atten}} = \sum_{u=1}^{L} p(u) \tilde a^{(u)},
\]
where $p(u){\ge}0~\forall u$ is a probability distribution (i.e., $\sum_{u=1}^{L}p(u)=1$) over the audio components. The probability distribution takes the form:
\[
p(u)\propto\exp\left(\alpha_l \theta_l(u) + \alpha_c \theta_c(u)\right).
\]
The local potential is $\theta_l(u)=v_l^{\top}\operatorname{relu}(V_la_u)$, and the cross potential between the audio components is:
\[
\theta_c(u) = \sum_{i=1}^{L} \left(\left(\frac{W_1\tilde a^{(u)}}{||W_1\tilde a^{(u)}||}\right)^\top\left(\frac{W_2\tilde a^{(i)}}{||W_2\tilde a^{(i)}||}\right)\right).
\]
The trainable parameters are (i) ${V_l, W_1, W_2}$, which re-embed the data to tune the attention, (ii) $v_l\in\mathbb{R}^{(L\cdot H\cdot d_t)\times 1}$  that scores the sound component (iii) ${\alpha_l, \alpha_c}$ that calibrates the local and cross potentials.
The attention mechanism enables learning the significance of the audio components.

\noindent{\textbf{Text-to-video.}} Lastly, we leverage a pre-trained latent diffusion text-to-video model to learn the aforementioned temporal audio tokens, $c = \{c^{(i)}\}_{i=1}^{L}$. 

Diffusion models are a family of generative models designed to learn the data distribution $p(x)$. This is done by learning the reverse Markov process of length $T$. Given a timestamp $t \in [0, 1]$, the denoising function $\epsilon_{\theta}: \R^d \rightarrow \R^d$ learns to predict a clean version of the perturbed $x_t$ from the training distribution. The generative process can be conditioned on a given input, i.e., modeling $p(x|y)$ where $y$ is a condition vector. In that case, the objective function is $\mathcal{L}_{\text{CLDM}} \triangleq$,
\[
    \E_{(v, a)\sim S, t\sim U(0, 1),\epsilon\sim\mathcal{N}(0, I)}\left[\norm{\epsilon - \epsilon_{\theta}(f\left(v_t, c\right), t)}_2^2\right],\label{eq:cldm}
\]
where each video frame, $v^{(i)}$, is conditioned on a dedicated condition vector $c^{(i)}$. 

Specifically, in this work, we set $\epsilon_{\theta}$ to be a state-of-the-art text-to-video model, ModelScope, which is comprised of a 3D-UNet integrated with a temporal attention layer as outlined in~\citet{wang2023modelscope}. ModelScope was trained on $\sim$10M text-video pairs and $\sim$2B text-image pairs~\cite{wang2023modelscope}. Notice the proposed framework is not limited to ModelScope and can be used over any differentiable text-to-video model.

\noindent{\textbf{Model optimization.}} We optimize the AudioMapper and the attentive pooling layer only and backpropagate gradients through $\epsilon_{\theta}$ while keeping its parameters unchanged. Optimization minimizes the loss $\mathcal{L}_{\text{CLDM}}$ for reconstructing a frame $v^{(i)}$ conditioned on $c^{(i)}$ (see Equation~\eqref{eq:cldm}), with an added weight decay regularization for the encoded \model. Overall, we optimize the following loss function: 
\begin{equation}
\mathcal{L} = \mathcal{L}_{\text{CLDM}} + \frac{\lambda_{l_1}}{L}\sum_{u=1}^{\log{L}} \tilde a^{(u)},
\end{equation}
where $\lambda_{l_1}$ is a trade-off hyper-parameter between the loss term and the regularization.
\section{Evaluation metrics}
We evaluate our method on three main axes: video quality and diversity, audio-video alignment, and a human study. 

\noindent{\textbf{Video quality and diversity.}} We report standard evaluation metrics in the domain of video generation for assessing quality and diversity. We utilize the following metrics: (i) Frechet Video Distance (FVD) metric, which quantifies the visual disparity between feature embeddings extracted from generated and reference videos \cite{unterthiner2019accurate} and is used to assess quality and diversity; (ii) Inception Score (IS), which is computed with a trained C3D model (\citet{tran2015learning}) on UCF-101~\cite{soomro2012ucf101} and assesses video quality.

\noindent{\textbf{Audio-video alignment.}}  We distinguish between two types of audio-video alignment: (i) Semantic (or global) alignment, in which the semantic class (e.g., playing drums) of the input audio is depicted by the output video (e.g., a video of people playing drums). To this end, we consider the CLIP Similarity (CLIPSIM) metric~\cite{wu2021godiva}, which gauges the alignment between generated video content and its corresponding audio label; (ii) Temporal alignment, in which we consider if each input audio segment is synchronized with its corresponding generated video segment. To measure this type of alignment, we introduce a novel evaluation metric. 

The new metric is based on detecting energy peaks in both modalities separately and measuring their alignment. 
The premise behind this metric is that fast temporal energy changes in the audio signal often correspond to an object movement producing this sound. For instance, consider an audio waveform of fireworks. A successful audio-video temporal alignment would ensure that the video frames portraying the fireworks exhibit a noticeable change synchronously. Conversely, when the video exhibits a significant change, a corresponding peak should be observed in the audio waveform at that precise moment. 

Our audio-video alignment metric operates as follows. We first detect candidate alignment points by considering each modality separately. We detect audio peaks using an Onset Detection algorithm~\cite{Boeck2013}, pinpointing instances of heightened auditory intensity. To detect the changes within the video, we calculate the mean of the Optical Flow~\cite{HORN1981185} magnitude for each frame and identify rapid changes over time. Then, for each peak in one modality, we validate whether a pick was also detected in the other modality within a three-frame temporal window and vise-verse. 

\begin{figure}[t!]
    \centering
    \includegraphics[width=\linewidth]{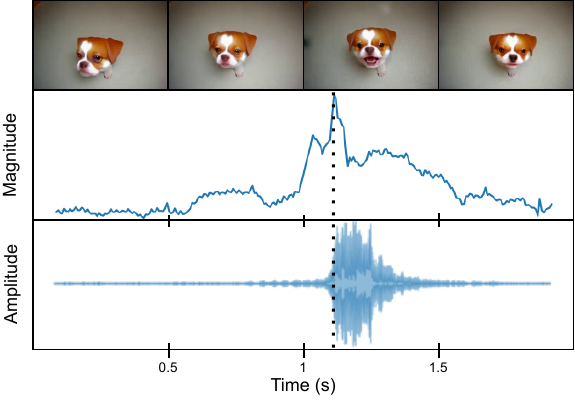}
    \caption{Audio-Video alignment metric illustration. The first row presents four frames from a generated video featuring a dog. The second row depicts the mean magnitude of optical flow for each frame, capturing video changes. The bottom row shows the amplitude of the audio waveform. The vertical line in the middle and the bottom graphs marks the onset of the waveform, while the peak of video change is also indicated.}
    \label{fig:audio_video_alignment}
\end{figure}

Finally, we normalize by the  number of peaks to derive the alignment score ranging between zero and one. Such a metric reflects the model's proficiency in synchronizing audio and video. More formally, given $\mathcal{A}$ and $\mathcal{V}$, audio and video peaks were obtained from the onset detection algorithms and optical flow, respectively. The alignment score is defined as: 
\[
    \text{AV-Align} = \frac{1}{2|\mathcal{A} \cup \mathcal{V}|} \left(\sum_{a\in\mathcal{A}} \indicator[a\in \mathcal{V}] + \sum_{v\in\mathcal{V}} \indicator[v\in \mathcal{A}]\right),
\]
where we consider a valid peak if placed within a window of three frames in the other modality. The above metric can be interpreted as the Intersection-over-Union metric. 

To facilitate comprehension, Figure \ref{fig:audio_video_alignment} illustrates the alignment process visually, depicting audio peaks and corresponding video changes, emphasizing the interplay between the auditory and visual domains.

\noindent{\textbf{Human study.}} 
We perform Mean Opinion Scores (MOS) experiments considering both quality and audio-video alignment. In this setup, human raters are presented with several short video samples and are instructed to evaluate their quality and alignment on a scale between 1--5 with increments of 1.0.  Specifically, we ask raters to evaluate the videos considering overall quality, global alignment to the audio file, and local alignment between the visual and sound of the video files. We evaluate 20 videos per method and enforce ten raters per sample. The full questionnaire we asked the raters can be found in the supplemental material.

\section{Experimental setup}
\noindent{\textbf{Implementation details.}} The proposed method contains $\sim$35M trainable parameters. We optimized the model using two A6000 GPUs for 10K iterations. We use AdamW optimizer with learning rate of 1e-05 using constant learning rate scheduler. Each batch comprises 8 videos with 24 frames per video, sampled randomly for one-second granularity. To enhance training efficiency and mitigate memory consumption, we integrated gradient checkpointing into the training process of the 3D U-net architecture. 
Code and pre-trained models will be publicly available upon acceptance.

\begin{table}[t!]
\centering\small
\setlength{\tabcolsep}{3pt}
\resizebox{.99\columnwidth}{!}{
\begin{tabular}{lcccc}
\toprule
Model & FVD ($\downarrow$) & CLIPSIM ($\uparrow$) & IS ($\uparrow$) & AV-Align ($\uparrow$) \\
\midrule
&\multicolumn{4}{c}{VGGSound}\\
\cmidrule{2-5}
ModelScope Text2Vid & \textbf{801} & \textbf{0.69} & \textbf{15.55} & 0.27\\
ModelScope Random & 1023 & 0.47 & 6.32 & 0.26\\
\midrule
Ours
& 923 & 0.57 & 11.04 & \textbf{0.35}\\
\midrule
&\multicolumn{4}{c}{AudioSet-Drums}\\
\cmidrule{2-5}
TATS & 303 & 0.69 & 2.10 & 0.28\\
\midrule
Ours
& \textbf{299} & \textbf{0.70} & \textbf{2.78} & \textbf{0.61}\\
\midrule
&\multicolumn{4}{c}{Landscape}\\
\cmidrule{2-5}
MM-Diffusion
& 922 & 0.53 & 2.85 & 0.41\\
\midrule
Ours
& \textbf{784} & \textbf{0.57} & \textbf{4.49} & \textbf{0.54}\\
\bottomrule
\end{tabular}}
\caption{Automatic video generation results. We report FVD, CLIPSIM, IS, and Alignment (`align') scores for both the proposed method (Ours) and the baselines. For a fair comparison, we compare our method to TATS~\cite{ge2022long} and to MM-Diffusion~\cite{ruan2023mm}  using the benchmarks reported by the authors in the original paper.}\label{tab:mainres}
\end{table}

\noindent{\textbf{Datasets.}} We utilize the VGGSound dataset \cite{chen2020vggsound}, derived from YouTube videos containing $\sim$180K clips of 10 seconds duration, annotated across 309 classes. To enhance data quality, we filtered $\sim$60K videos in which audio-video alignment is weak. During this filtering procedure, we utilized a pre-trained audio classifier to categorize sound events present in each clip. Simultaneously, a pre-trained image classifier was employed to classify the middle frame of every video clip. We then computed the CLIP~\cite{radford2021learning} score by comparing the predicted labels from both classifiers. Then, filtering is done by removing videos that do not pass a pre-defined threshold. 

Additionally, to have a fair comparison with prior work, we experimented with two additional datasets. (i) The \emph{Landscape} dataset \cite{lee2022sound}, which contains 928 nature videos divided into 10-second clips, covering nine distinct scenes; (ii) The \emph{AudioSet-Drum} dataset~\cite{gemmeke2017audio}, contains $\sim$7k videos of drumming. We used the same split as proposed by~\citet{ge2022long}, where $\sim$6k is used as the training set while the rest serves as a test set.

\noindent{\textbf{Baselines.}} We compare the proposed method to previous state-of-the-art models generating videos conditioned on audio inputs. \citet{ge2022long} proposed Time Sensitive Transformer (TATS) model, which projects audio latent embeddings onto video embeddings, enabling cross-modal alignment. \citet{ruan2023mm} recently proposed MM-Diffusion, which employs coupled denoising auto-encoders to generate joint audio and video content. Each of the above-mentioned baselines, i.e., TATS and MM-Diffusion, were originally evaluated using different benchmarks, i.e., AudioSet-Drums and Landscape, respectively. For a fair comparison, we evaluate the proposed method using each of the datasets suggested in the original papers. 

Moreover, we consider two naive baselines based on text-to-video models. In the first one, we generate videos from text description and retrieve random audio from the training set which corresponds to the same class as the generated video, denoted as \emph{ModelScope Text-To-Video}. For the second one, denoted as \emph{ModelScope Random}, we generate videos unconditionally (i.e., without any specific textual conditions), and match it with a random audio segment. For both baselines, we use the pre-trained publicly available zeroscope-v2 model~\footnote{we use the zeroscope-v2 576w as can be found in the following link: \url{https://huggingface.co/cerspense/zeroscope_v2_576w}}. 
\begin{figure}[t!]
    \centering
    \includegraphics[width=\linewidth]{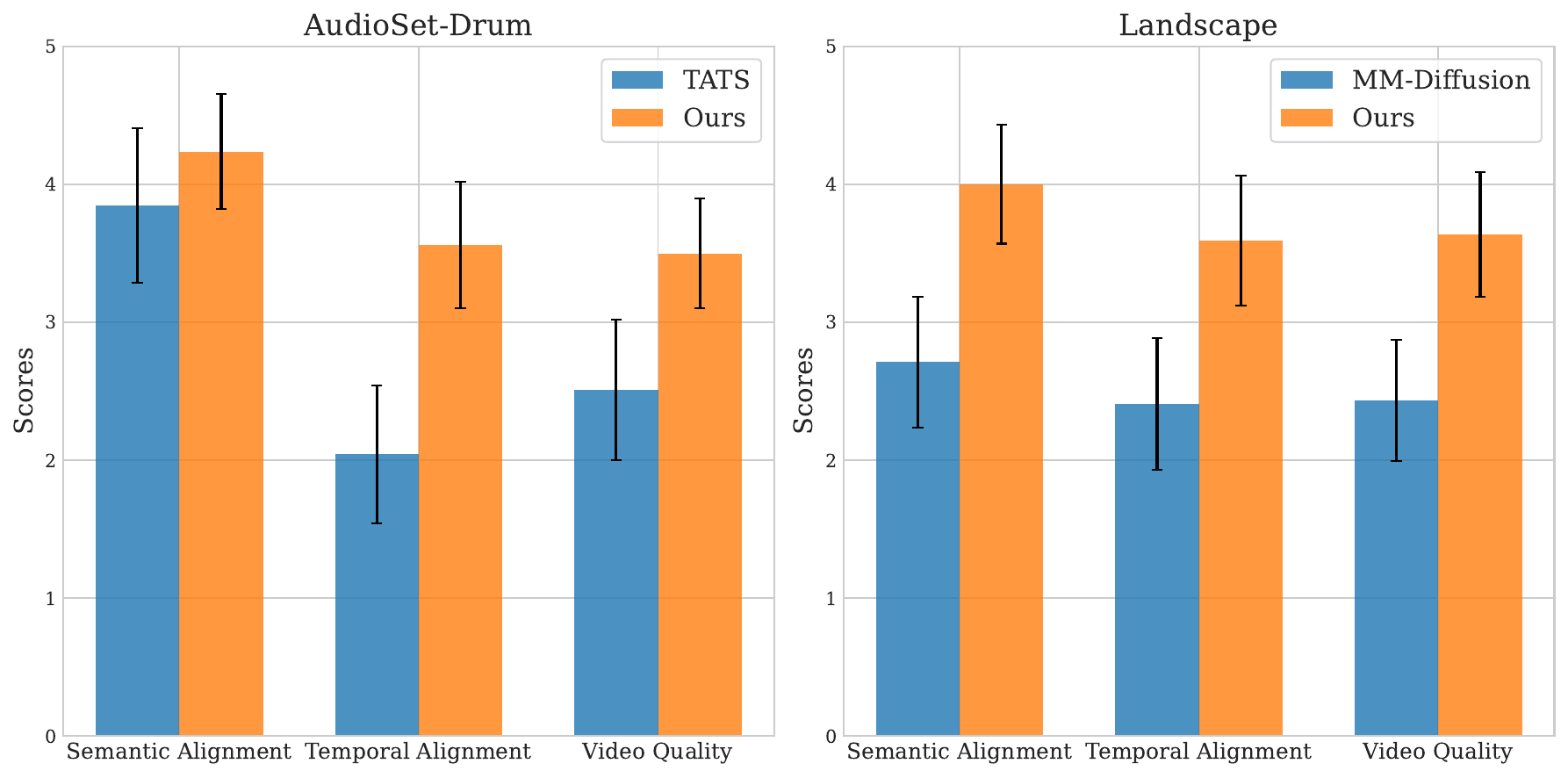}
    \caption{Human study. We consider the MOS score for three metrics: (i). \textit{Semantic alignment}, where we ask users to rate how well the video matches the input audio semantic label, (ii). \textit{Temporal alignment}, where we ask users to rate how well each input audio segment is aligned with the generated video segments, and (iii) \textit{Video quality}, where we ask users to rate the generated video quality. On the LHS, we consider video models trained on AudioSet-Drum, and on the RHS, we consider video models trained on Landscape.} \label{fig:human_eval}
\end{figure}

\section{Results}
We start by presenting results for audio-to-video generation considering both objective metrics presented above and human study. Next, we empirically demonstrate how the proposed method can be used to generate videos conditioned on both text and audio modalities, thus enhancing text-to-video generations. Lastly, we conduct an ablation study to understand better the effect of our audio conditioning technique on generation quality and alignment. Visual results are provided in the supplementary. 

\subsection{Audio-to-video generation}
\noindent{\textbf{Objective evaluation.}} As can be seen in Tab.~\ref{tab:mainres}, our method outperforms the baselines on all metrics for the AudioSet-Drums and Landscape datasets. Specifically, our method improves both the quality of the generated videos (FVD and IS scores) together with the audio-video alignment (AV-Align and CLIPSIM scores). As expected, the gap between the methods is larger when considering the alignment scores.

Notice the alignment scores changed significantly when considering different benchmarks. Sound events can also be produced by objects not seen in the video; this is especially noticeable in the VGGSound benchmark, in which the AV-Align score of the original videos is $0.51$. 

Next, we compare our method to the original ModelScope model, both text-condition (ModelScope Text2Vid) and unconditionally (ModelScope Random). As we do not modify the model, we consider the text-condition setup as a top-line in terms of video quality metrics. Recall the audio in both models is retrieved from our training set, using either the video class for ModelScope Text2Vid or randomly ModelScope Random. As expected, our model outperforms ModelScope Random considering all metrics. The ModelScope Text2Vid is superior to our model for video quality. However, when considering audio-video alignment, our method is significantly better.

\noindent{\textbf{Human study.}} We present results using a human study considering both video quality and alignment (both semantic and temporal). Results are depicted in Figure~\ref{fig:human_eval}. As can be seen for both the AudioSet-Drum and Landscape datasets, users found our videos significantly more temporally aligned. For semantic alignment, our method improves on both TATS and MM-Diffusion, with a significant gap to MM-Diffusion on the Landscape dataset. Finally, on video quality, users found our videos significantly superior. 

\subsection{Joint audio-text to video generation}

Utilizing text and audio together to guide generation involves adding text tokens for conditioning. In Tab.~\ref{tab:text_audio_res}, we show results using ``A video of $<$class$>$'' for text conditioning and ``A video of $<$TemporalAudioAtokens$>$ $<$class$>$'' for Text+Audio. Combining text and audio conditioning outperforms audio-only in all metrics, especially FVD. Text-only provides the highest video quality but lacks alignment.

\begin{figure}[t!]
    \centering
    \includegraphics[width=\linewidth]{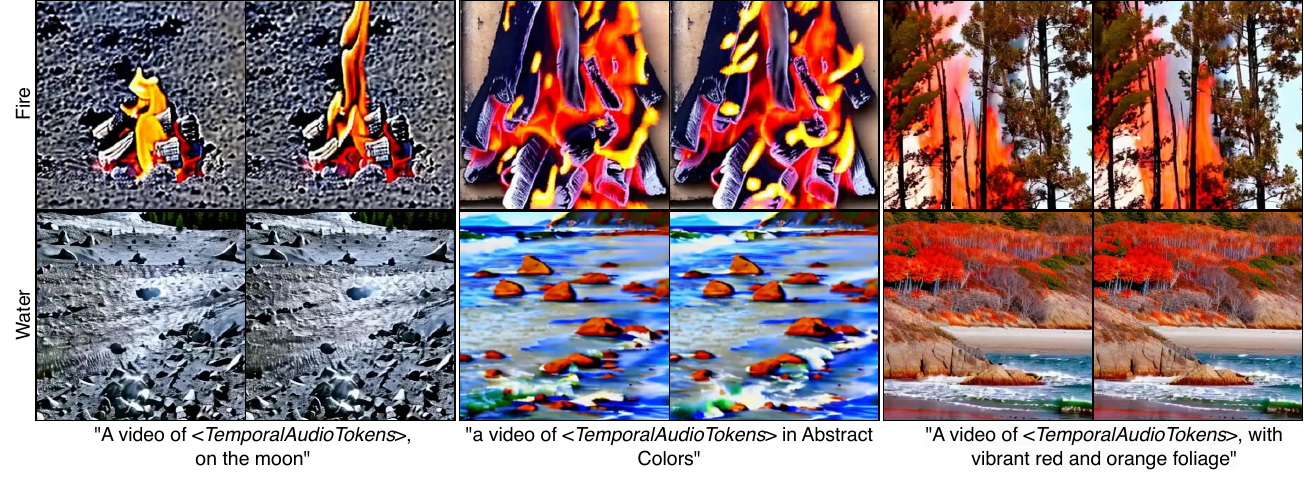}
    \caption{Examples of added text tokens for altering the output video. We show results for fire and flowing water audio.}
    \label{fig:audiotext}
\end{figure}

In Fig.~\ref{fig:audiotext}, we present how we merge text tokens to temporal audio tokens, which enables style manipulation. For example, for the sound of a river, we can depict it flowing over the moon by using the prompt ``on the moon''.

\begin{table}[t!]
\centering
\resizebox{.99\columnwidth}{!}{
\begin{tabular}{lcccc}
\toprule
Cond. & FVD ($\downarrow$) & CLIPSIM ($\uparrow$) & IS ($\uparrow$) & AV-Align ($\uparrow$) \\
\midrule
Text & \textbf{801} & \textbf{0.69} & \textbf{15.55} & 0.27\\
\cmidrule{2-5}
Audio & 923 & 0.57 & 11.04 & 0.35\\
\cmidrule{2-5}
Text+Audio & 859 & 0.58 & 11.66 & \textbf{0.36} \\
\bottomrule
\end{tabular}}
\caption{Results of the proposed method using different modalities as conditioning. We report results for Text, Audio, and Text+Audio modalities as model conditioning.}\label{tab:text_audio_res}
\end{table}

\subsection{Ablation study}
Recall our method consists of using context windows of varying sizes to capture a local-to-global context of the input audio. In Tab.~\ref{tab:abl_res}, we assess the effect of using different windows of size $K \in \{1, 2, 3, 4\}$ denoted as win. (K-res.). Note in practice, the window size is determined by $\log K$; we use $K$ for readability. Using only the local context window ($K=1$) results in a good alignment. As we increase the global context (i.e., increasing $K$), the video quality is improved while the alignment scores are comparable. 

 We additionally consider a single audio conditioning vector (vec) by averaging all the audio components. Despite high video quality scores, the absence of local temporal information results in a notably worse AV-Align score.

\begin{table}[t!]
\centering
\resizebox{.99\columnwidth}{!}{
\begin{tabular}{lcccc}
\toprule
Cond. & FVD ($\downarrow$) & CLIPSIM ($\uparrow$) & IS ($\uparrow$) & AV-Align ($\uparrow$) \\
\midrule
vec. & 948 & 0.57 & 10.12 & 0.29\\
\cmidrule{2-5}
win. (1-res.)  & 998 & 0.56 & 9.22 & \textbf{0.36}\\
\cmidrule{2-5}
win. (2-res.) & 965 & 0.56 & 9.87 & 0.35\\
\cmidrule{2-5}
win. (3-res.) & 972 & 0.56 & 10.01 & 0.34\\
\cmidrule{2-5}
win. (4-res.) & 950 & 0.56 & 10.13 & 0.35\\
\cmidrule{2-5}
win. (5-res.) & \textbf{923} & \textbf{0.57} & \textbf{11.04} & 0.35\\
\bottomrule
\end{tabular}}
\caption{An ablation study exploring the different audio conditioning. We report FVD, CLIPSIM, IS, and Alignment scores on VGGSound \cite{chen2020vggsound} considering single-vector conditioning (vec.), time-dependent condition using one window size (win. (1-res.), and different windows of size $k$ (win. ($k$-res.)).}\label{tab:abl_res}
\end{table}

\section{Limitations}
As the proposed method leverages a pre-trained text-to-video model, the adaptation process between text to audio tokens requires mapping between both latent representations. As both modalities operate at different levels of granularity, it is unclear whether such mapping holds all the relevant information in the audio space. Moreover, at the moment, our method generates relatively short video segments since the temporal conditioning is limited to 24 frames due to hardware limitations. 

Lastly, while audio can indeed convey information about a visual scene, discrepancies can also arise between the two modalities. For example, a video might depict a dog in a car while the accompanying audio only features a radio playing. 
This disparity is particularly noticeable in the context of shorter videos. Such mismatch imposes a general limitation of the domain at large, not specifically to our method. 

\section{Conclusion}
In this work, we introduced a state-of-the-art audio-to-video generation model capable of generating diverse and realistic videos aligned to input audio samples. By learning a lightweight adapter to map between the input audio representation and a text-based representation, video generation can be conditioned not only on audio but also on text, enabling, for the first time, the generation of audio aligned to both input audio and text samples. To better capture both local and global context around each frame, we consider an expanding context window technique. 
We validate our approach extensively,  demonstrating significant semantic diversity of audio-video samples, and further propose a novel evaluation
metric (AV-Align) to assess the temporal alignment of the input audio and generated video. 
For future work, we are excited to explore how further modalities, such as depth, images, or IMU can be used jointly with audio and text as further conditions by which video can be generated. 


 \newpage
\bibliography{aaai24}

\end{document}